\documentclass[nohyperref]{article}

\usepackage{microtype}
\usepackage{graphicx}
\usepackage{subfigure}
\usepackage{booktabs} 

\usepackage{hyperref}

\usepackage[accepted]{icml2022}

\usepackage{amsmath}
\usepackage{amssymb}
\usepackage{mathtools}
\usepackage{amsthm}

\usepackage[section]{placeins}

\usepackage[capitalize,noabbrev]{cleveref}

\theoremstyle{plain}

\theoremstyle{definition}

\theoremstyle{remark}

\usepackage[textsize=tiny]{todonotes}

\icmltitlerunning{Dimensionality Reduction Meets Message Passing for Graph Node Embeddings}

\begin{document}

\twocolumn[
\icmltitle{Dimensionality Reduction Meets Message Passing for Graph Node Embeddings}



\begin{icmlauthorlist}
\icmlauthor{Krzysztof Sadowski}{compIntelPL}
\icmlauthor{Michał Szarmach}{compIntelPL}
\icmlauthor{Eddie Mattia}{compIntelUS}

\end{icmlauthorlist}

\icmlaffiliation{compIntelPL}{Intel}
\icmlaffiliation{compIntelUS}{SigOpt (Intel)}

\icmlcorrespondingauthor{Krzysztof Sadowski}{krzysztof.sadowski@intel.com}
\icmlcorrespondingauthor{Eddie Mattia}{eddie.mattia@intel.com}
\icmlcorrespondingauthor{Michał Szarmach}{michal.szarmach@intel.com}

\icmlkeywords{Machine Learning, ICML}

\vskip 0.3in
]



\printAffiliationsAndNotice{}  

\begin{abstract}
Graph Neural Networks (GNNs) have become a popular approach for various applications, ranging from social network analysis to modeling chemical properties of molecules.
While GNNs often show remarkable performance on public datasets, they can struggle to learn long-range dependencies in the data due to over-smoothing and over-squashing tendencies.
To alleviate this challenge, we propose PCAPass, a method which combines Principal Component Analysis (PCA) and message passing for generating node embeddings in an unsupervised manner and leverages gradient boosted decision trees for classification tasks.
We show empirically that this approach provides competitive performance compared to popular GNNs on node classification benchmarks, while gathering information from longer distance neighborhoods.
Our research demonstrates that applying dimensionality reduction with message passing and skip connections is a promising mechanism for aggregating long-range dependencies in graph structured data.
\end{abstract}

\label{submission}

\section{Introduction}
Deep learning on graphs has become a common building block~\cite{zhang2020deep,gnnsurvey2021} for applications like social network analysis~\cite{li2021relevanceaware}, recommendation systems~\cite{gao2021graph}, fraud detection mechanisms~\cite{wang2021review} and predicting chemical properties~\cite{li2021graph}. 
Unlike previous approaches that involved converting a graph into a sequence of tensors, GNNs use its structure to learn more expressive representations.~\cite{hamilton2018representation,zhou2021graph}.
\begin{figure}[H]
    \vskip 0.2in
    \begin{center}
    \centerline{\includegraphics[width=\columnwidth]{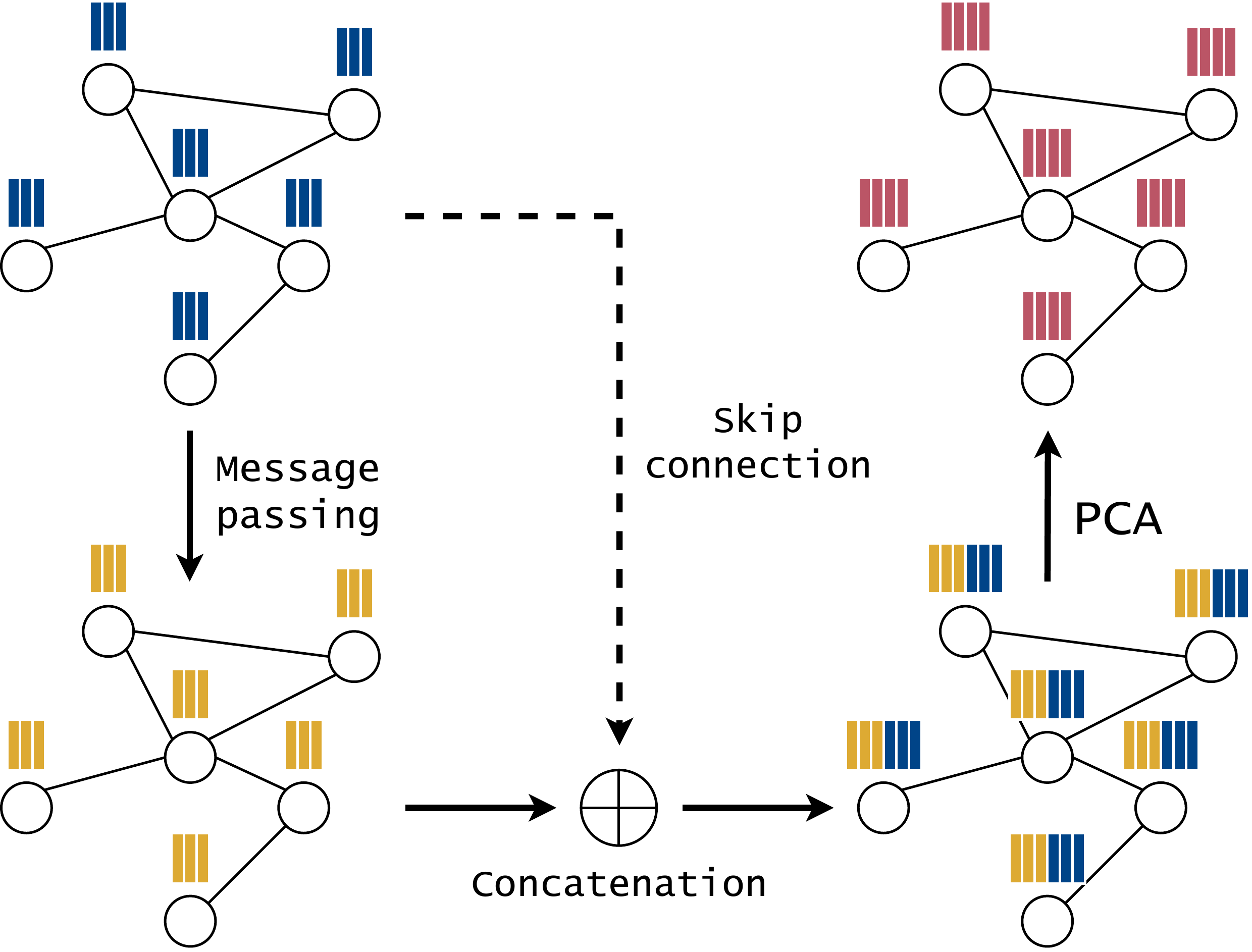}}
    \caption{Illustration of the process performed in the PCAPass layer. Our method keeps the meaningful properties from the aggregated neighborhood and node embeddings from the previous algorithm step by using dimensionality reduction.}
    \label{fig:pcapass_schema1}
    \end{center}
    \vskip -0.2in
\end{figure}
Most GNN architectures distinguish themselves from other neural network families by the concept of messaging passing~\cite{gilmer2017neural}. 
The essence of this operation is that the node collects messages from its neighbors, i.e. other nodes connected by the edge. 
The entire process can be described as follows:
\begin{equation}
m^{k}_v = \sum_{u\epsilon\mathcal{N(\upsilon)}} M_k(h^{k-1}_u,h^{k-1}_v,e_{uv})
\label{eq:mpassing}
\end{equation}
\begin{equation}
h^{k}_v = U_k(h^{k-1}_v,m^{k}_v),
\label{eq:update}
\end{equation}
where $h$ is node feature embedding, $k$ is an algorithm step and $N$ denotes nodes in graph. 
Messages are computed as a message function $M$ of the source node $u$, destination node $v$ , and the edge connecting them $e$ (Equation~\ref{eq:mpassing}). 
Then, aggregated messages $m$ by permutation invariant function are used to create new embedding of destination node by update function $U$ (Equation~\ref{eq:update}). 
Repeating this process $k$ times causes the node to incrementally receive information from further $k$-hop neighborhoods.

As the receptive field is enlarged, the number of neighbors increases exponentially leading to a problem known as neighborhood explosion which leads to high memory utilization. 
For tasks such as predicting molecular properties, where GNNs operate on many small graphs, it is possible to batch them to fit into memory. 
In the case of training on one large graph, such as a social network, many solutions related to sampling of neighbors or graph partitioning were created \cite{pingsage2018,cluster2019,hamilton2018representation}. 
However, this process can potentially introduce bias to the model and is computationally expensive as it is typically performed for each layer of the neural network.

Stacking multiple layers of message passing is necessary to train long-range dependencies in a graph but can also lead to issues known as over-smoothing and over-squashing. 
Due to the aggregation function in message passing as more and more layers are added, adjacent nodes begin to have increasingly similar representations~\cite{li2018deeper}.
We define the problem of over-smoothing by the difficulty of the predictive tasks in distinguishing between nodes. The term over-squashing has been introduced to describe a similar problem with GNNs not propagating long-range graph signals. As the information is aggregated from an exponentially growing number of neighbors, it is propagated through a constant size vector, leading to a bottleneck like that found in Recurrent Neural Networks~\cite{alon2021bottleneck}.

Typically, GNNs use a multilayer perceptron (MLP) to make predictions.
In the case of our method, we decide to use the gradient boosted decision trees classifier (GBDT) implemented with the XGBoost algorithm \cite{xgboost2016}.
Gradient boosting is a widely used machine learning technique.
An ensemble of weak models, usually decision trees, is used to create a stronger predictive model.
Like other boosting methods, GBDT is built iteratively but allows optimization of any differentiable loss function.

In this paper we present PCAPass (Figure \ref{fig:pcapass_schema1}), a node embedding method that addresses the problems of learning information from distant neighborhoods in graph structured data. 
Our \textbf{contributions} are as follows:

\begin{itemize}
\item We introduce resistant to over-smoothing method for generating node embeddings in unsupervised manner that leverages graph structure and input features. We empirically prove that our method benefits over message passing itself in gathering information from long-range signals
\item By using message passing in preprocessing our method is more computationally efficient than GNNs which use this process during training. Additionally, therefore, it can be used by the GBDT classifier to make use of the graph structure or other GNNs using message passing in preprocessing.
Our experiments show that PCAPass gives a better generalization of XGBoost to unseen data than message passing itself.
\item We empirically demonstrate that PCAPass combined with the XGBoost classifier gives competitive results compared to popular GNNs on three benchmark datasets: ogbn-arxiv, ogbn-products and Reddit.
\end{itemize}

\section{Related Work}
When training GNNs, the goal is to make predictions at the node, edge, or graph level.
The structure of the graph on which the model operates determines the patterns that must be learned.
In order to aggregate long-range signals efficiently, many solutions have been developed that address the problem of over-smoothing and over-squashing.

\subsection{Learning Distant Relationships}
Recently, several solutions have been developed to avoid over-smoothing using the skip connection mechanism known from Convolutional Neural Networks \cite{li2019deepgcns, li2020deepergcn, gong2020geometrically, xu2021optimization}.
Instead of summing node embeddings from the previous layer along with the aggregated information from neighbors, GCNII uses the initial residual connection to propagate the input features across all layers \cite{chen2020simple}.
JKNet leverages this mechanism to combine the node embeddings from the various message passing steps to get better structure aware representations \cite{xu2018representation}.
These solutions use multiple layers of message passing while training, and therefore consume a lot of memory.
To improve the scalability of deep GNNs, RevGNN \cite{li2021training} extends this mechanism using reversible residual connections and applies additional approaches such as group convolutions and weight tying. This model achieves a state-of-the-art result on the ogbn-proteins dataset.

Over-squashing is implicitly resolved through solutions that employ an attention mechanism. 
Self-attentional layers used in GAT \cite{velickovic2018graph} and GATv2 \cite{brody2021attentive} alleviate over-squashing by weighting messages can ignore irrelevant edges, which means that fewer of them have to pass through the bottleneck \cite{alon2021bottleneck}. 
The initial residual approach has been combined with the concept of attention in FDGATII, while mitigating both of the mentioned problems related to learning long-range dependencies \cite{kulatilleke2021fdgatii}. 
Another recent work discusses the relation between the geometric concept of Ricci Curvature
and over-squashing and proposes a graph rewiring method to alleviate this problem \cite{topping2021understanding}.

\subsection{Message Passing in Preprocessing}
To avoid high memory consumption and enable training without node sampling, SIGN uses message passing in preprocessing \cite{rossi20sign}. 
The success and scalability of this method have inspired other solutions to use this approach. 
The NARS authors used a similar mechanism to train the model on heterogeneous graphs, i.e. graphs with multiple relations \cite{yu2020scalable}. 
Receptive field attention introduced in GAMLP enable nodes to flexibly use aggregated messages during precomputation from different neighborhood sizes \cite{zhang2021graph}. 
The methods using neighborhood aggregation in preprocessing yield state-of-the-art results on Open Graph Benchmark node classification datasets \cite{hu2021open}.

\subsection{Gradient Boosted Decision Trees on Graph Data}
Motivated by the observation that GBDT are usually better at making predictions from heterogeneous features than other machine learning methods, BGNN authors use this method for graph data \cite{ivanov2021boost}. 
This solution can be viewed as an inversion of ours because instead of using a graph structure to prepare node embeddings, it uses GBDT to process the features that are passed to a GNN. 

\begin{figure}[H]
    \vskip 0.2in
    \begin{center}
    \centerline{\includegraphics[width=\columnwidth]{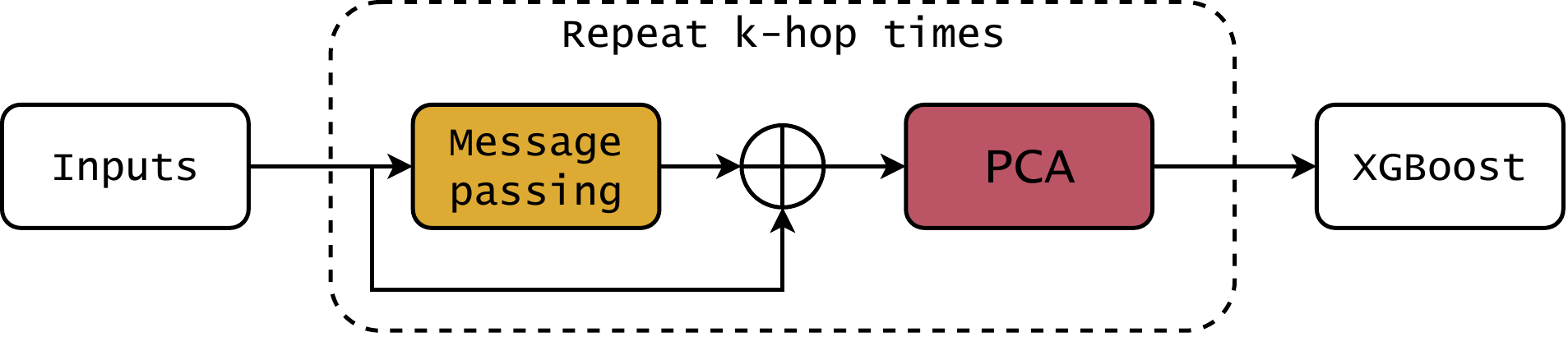}}
    \caption{End-to-end architecture of PCAPass method combined with XGBoost classifier. We use \textbf{ $\bigoplus$} to denote concatenation.}
    \label{fig:pcapass_pipeline}
    \end{center}
    \vskip -0.2in
\end{figure}
\section{PCAPass}
In this work we propose PCAPass, a method for generating node embeddings in an unsupervised manner. 
The key building block of the architecture is the Principal Component Analysis (PCA) dimensionality reduction process combined with the message passing mechanism. 
PCA is based on selecting only the first principal components to obtain lower-dimensional data while preserving as much of the data's variation as possible.
Drawing inspiration from the SIGN architecture, our method uses message passing in preprocessing.
The success of this model suggests that GNNs are using message passing to smooth node features over local neighborhoods rather than learning non-linear feature hierarchies. 
Moreover, giving up stacking message passing with MLP allows GBDT to leverage graph structure without any additional mechanism that requires gradient backpropagation. 
PCAPass method (Figure~\ref{fig:pcapass_pipeline}) can be described as follows:
\begin{equation}
h^k _\upsilon = AGG_k(h^{k-1}_u, \forall_u  \; \epsilon  \; \mathcal{N(\upsilon)})
\label{eq:agg}
\end{equation}
\begin{equation}
h^k_\mathcal{N(\upsilon)} = PCA(h_v^k \; \| \; h_v^{k-1}),
\label{eq:pca}
\end{equation}
where $AGG$ is local neighborhood feature aggregation function and $\|$ is concatenation. 
At each iteration, our algorithm gathers node information from local neighbors (Equation~\ref{eq:agg}). 
The architecture design allows the choice of any aggregation functions suitable to graph structure and features. 
After gathering information from each node's neighborhood, PCAPass uses skip connections to provide node embeddings from the previous step and concatenates both vectors (Equation~\ref{eq:pca}). 
The use of concatenation for skip connections instead of sum as usual is also seen in GraphSAGE.
At each iteration, nodes aggregate information from more distant neighbors, and the size of information used to compute new embeddings increases exponentially $(2^k)$.
We apply dimensionality reduction (Equation ~\ref{eq:pca}) at each step of the algorithm to overcome large memory consumption and high computational costs.
In addition, this process retains essential properties from aggregated neighborhood and of destination node state from the previous iteration.   
It is worth noting that PCAPass is not using node sampling or subgraph partitioning which could introduce bias into training procedure.

\section{Experiments}
We evaluate PCAPass performance on three benchmarks: predicting Amazon co-purchasing network categories in a multiclass configuration (ogbn-products), classifying scientific articles into different subject areas using the citation network (ogbn-arxiv), and classifying Reddit posts as belonging to different communities (Reddit). In all these experiments, we test the proposed method on the node classification task.
\begin{table}[H]
    \centering
    \caption{Classification accuracy on tested datasets. 
    PCAPass combined with XGBoost achieves the best results from the compared methods on ogbn-products and gives competitive results on the others.}
    \label{accuracy-table}
    \vskip 0.15in
    \begin{center}
    \begin{small}
    \begin{sc}
    \fontsize{6.5}{7}\selectfont
    \begin{tabular}{p{2.4cm} cccc}
    \toprule
    \textbf{model} & \textbf{ogbn-arxiv} & \textbf{ogbn-products} & \textbf{reddit} \\
    \midrule
    DepeerGCN&71.92$\pm$0.16&80.98$\pm$0.20&N/A\\
    GAT&71.59$\pm$0.38&79.45$\pm$0.57&N/A\\
    GraphSAGE&71.74$\pm$0.29&78.70$\pm$0.36&95.40$\pm$0.22\\
    SIGN&\textbf{71.95$\pm$0.06}&80.52$\pm$0.16&\textbf{96.80$\pm$0.00}\\
    \textbf{PCAPass + XGBoost}&71.87$\pm$0.03&\textbf{81.15$\pm$0.02}&96.26$\pm$0.02\\
    XGBoost&54.56$\pm$0.07&61.83$\pm$0.02&74.88$\pm$0.07\\
    \bottomrule
    \end{tabular}
    \end{sc}
    \end{small}
    \end{center}
    \vskip -0.1in
    \end{table}

\subsection{Baselines}
The effectiveness of our method is compared with the popular GNNs, that have a similar approach to the problems related to learning long-range signals.
DeeperGCN uses skip connections and graph normalization to alleviate over-smoothing. 
Similar to GraphSAGE, PCAPass combines aggregated messages and node embedding from the previous step of the algorithm by concatenation. 
GAT addresses the tendency of over-smoothing \cite{min2022scattering} and over-squashing \cite{alon2021bottleneck} with an attention mechanism that weights incoming messages. 
SIGN performs message passing in a preprocessing step that generates an individual node embedding for each $k$-hop step. 
Therefore, it requires \textit{k - 1} times more memory to store preprocessed input features than PCAPass with the same embedding size.

\subsection{Setup}
In all experiments, we use graphs with bidirectional edges and self-loops. Early stopping is used during classifier training with a patience of 10 with respect to the calculation of cross-entropy loss on the validation dataset split. We experimented with mean aggregation and the symmetric normalized adjacency matrix as aggregation rules. Empirically, we found that the mean aggregation is better suited to the PCAPass method on the tested datasets. An attempt to standardize the features at each $k$-hop step before applying the PCA process to the concatenated embeddings (Equation~\ref{eq:pca}) has shown that on the tested datasets this approach is not beneficial. The architectural and optimization hyperparameters were sampled using SigOpt\footnote{app.sigopt.com} for Bayesian optimization.

\begin{table}[H]
        \centering
        \caption{Number of $k$-hop neighborhoods aggregated during training. 
        PCAPass along with DeeperGCN that also uses skip connections have a larger receptive field than other methods.}
        \label{khop-table}
        \vskip 0.15in
        \begin{center}
        \begin{small}
        \begin{sc}
        \fontsize{6.5}{7}\selectfont
        \begin{tabular}{lccc}
        \toprule
        \textbf{model} & \textbf{ogbn-arxiv} & \phantom{1} \textbf{ogbn-products} & \phantom{1} \textbf{reddit} \\
        \midrule
        DeeperGCN&\phantom{1}\textbf{28}&\phantom{1}14&  \phantom{1}N/A\\
        GAT&\phantom{1}3&\phantom{1}3&\phantom{1}N/A\\
        GraphSAGE&\phantom{1}3&\phantom{1}3&\phantom{1}3\\
        SIGN&\phantom{1}5&\phantom{1}5&\phantom{1}5\\
        \textbf{PCAPass + XGBoost}&\phantom{1}13&\phantom{1}\textbf{24}&\phantom{1}\textbf{21}\\
        \bottomrule
        \end{tabular}
        \end{sc}
        \end{small}
        \end{center}
        \vskip -0.1in
        \end{table}

\subsection{Results}
Results of our experiments, performed over 10 different random seeds, are summarized in the Table~\ref{accuracy-table}.
PCAPass works best compared to other methods on ogbn-products, which can be viewed as a semi-supervised task as it contains 90\%  of nodes in the test data split. 
On the smaller ogbn-arxiv and Reddit graphs, our method outperforms GraphSAGE and GAT and provides competitive results to DeeperGCN and SIGN. 
Combining PCAPass with XGBoost gives results at least 15\% better than GBDT alone on all tested datasets.

Additionally, in Table~\ref{khop-table} we show how big is the receptive field of methods that use message passing.
GBDT classifier trained on PCAPass node embeddings leverages longer-range dependencies than other methods that are not using skip connections. DeeperGCN gathers information from more distant neighborhoods than PCAPass on ogbn-arxiv.

\subsection{Implementation}
In our method, message passing is implemented using Deep Graph Library~\cite{wang2020deep} with PyTorch~\cite{paszke2019pytorch} as backend. We use PCA from scikit-learn~\cite{pedregosa2018scikitlearn} with the Intel scikit-learn extension\footnote{https://github.com/intel/scikit-learn-intelex} to leverage software and hardware stack. Our choice for GBDT classifier is XGBoost, a popular open-source library which is designed to be flexible, portable, and efficient solution. All experiments were run on Intel(R) Xeon(R) Platinum 8375C CPU @ 2.90GHz with 128 GB RAM using Ubuntu 18.04 operating system.

\section{Analysis}
In early experiments, we tested the simple preprocessing message passing solution. 
Therefore, we decided to make additional analyzes to show our method in the context of generalization to unseen data and gathering information from a large receptive field.

\begin{figure}[H]
    \vskip 0.2in
    \begin{center}
    \centerline{\includegraphics[width=\columnwidth]{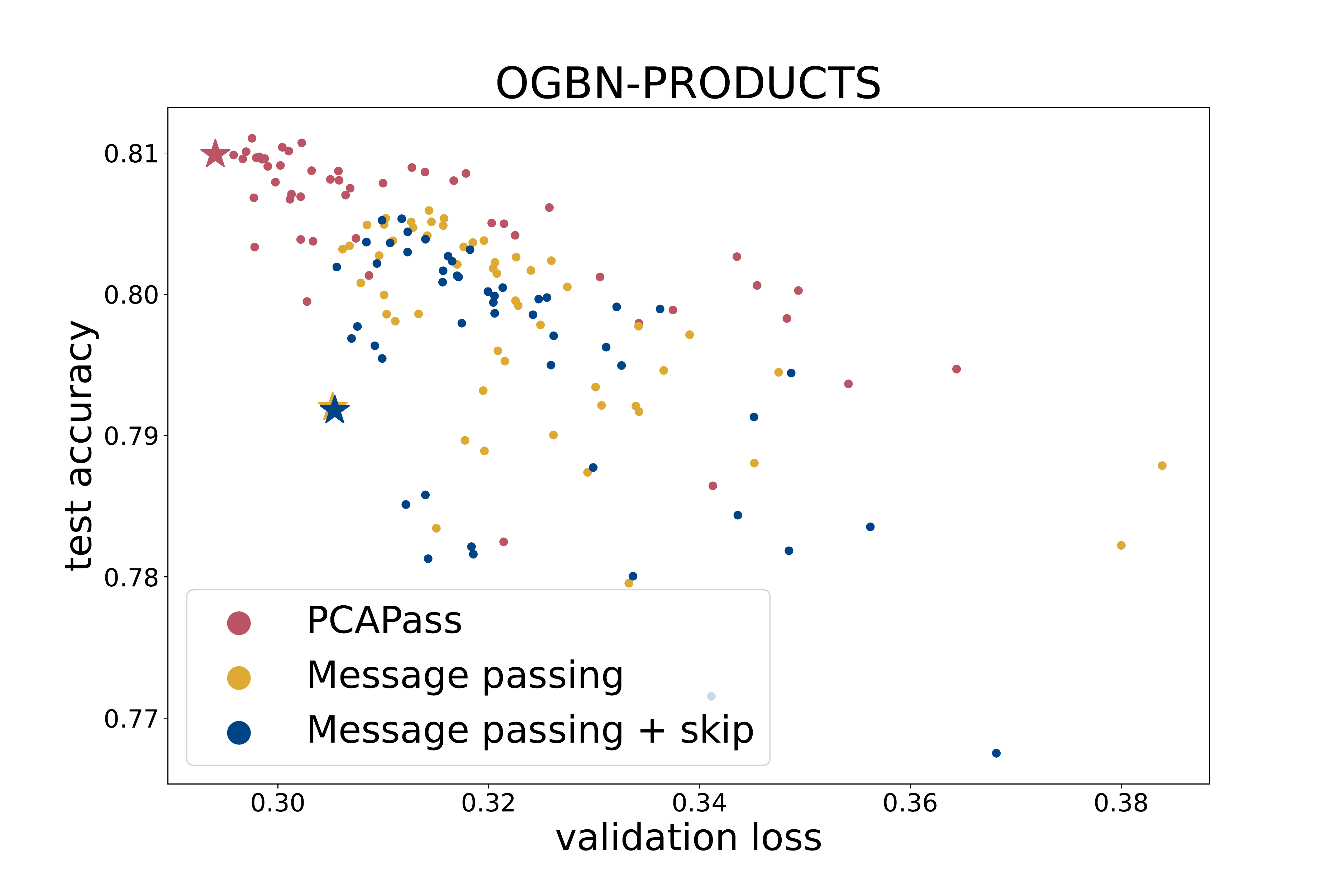}}
    \caption{XGBoost classifier performance trained on node embeddings during 50 runs of hyperparameter optimization. The best result in terms of validation loss is marked with a \textbf{\large $\star$} for the specific method.}
    \label{fig:pcapass_validation}
    \end{center}
    \vskip -0.2in
\end{figure}

\begin{figure*}[!hbt]
    \vskip 0.2in
    \begin{center}
    \centerline{\includegraphics[width=1.0\textwidth]{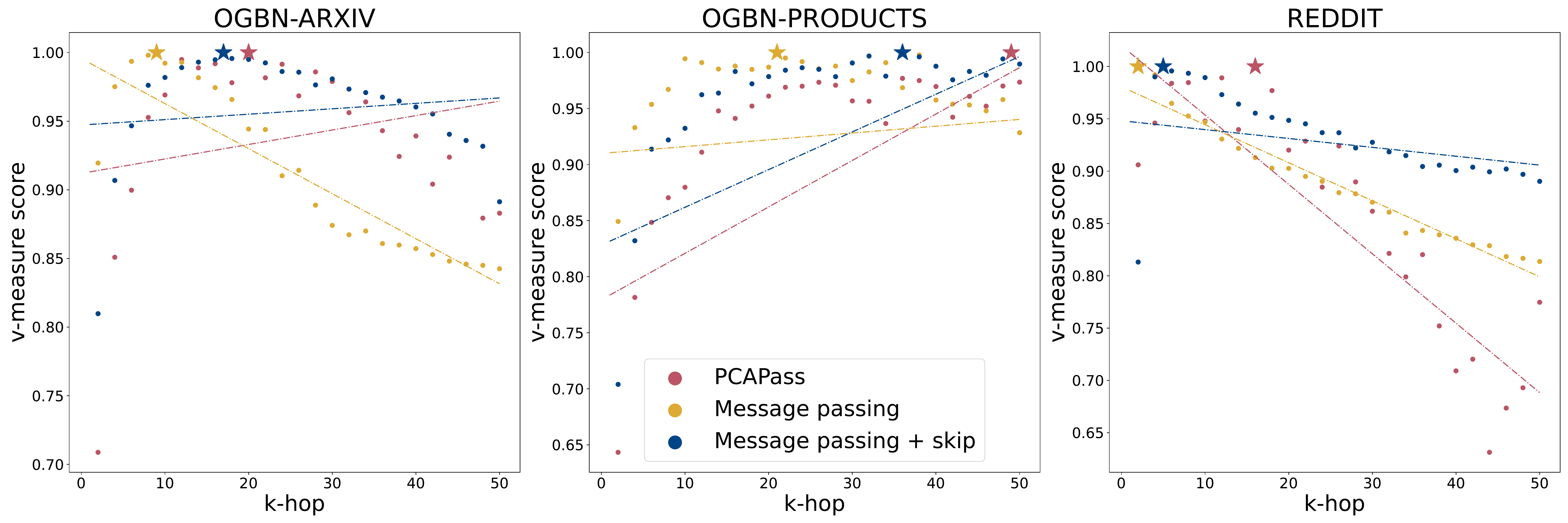}}
    \caption{Analysis of resistance to over-smoothing of node embeddings performed with k-means clustering. Normalized v-measure score per method shows how each of them behaves while changing the size of receptive field. The trend lines and best results denoted by \textcolor{black}{\large$\star$} indicate that PCAPass and message passing with skip connections are least affected by over-smoothing.}
    \label{fig:vms}
    \end{center}
    \vskip 0.2in
\end{figure*}

\subsection{Generalization to Unseen Data}
\begin{figure}[H]
    \vskip 0.2in
    \begin{center}
    \centerline{\includegraphics[width=\columnwidth]{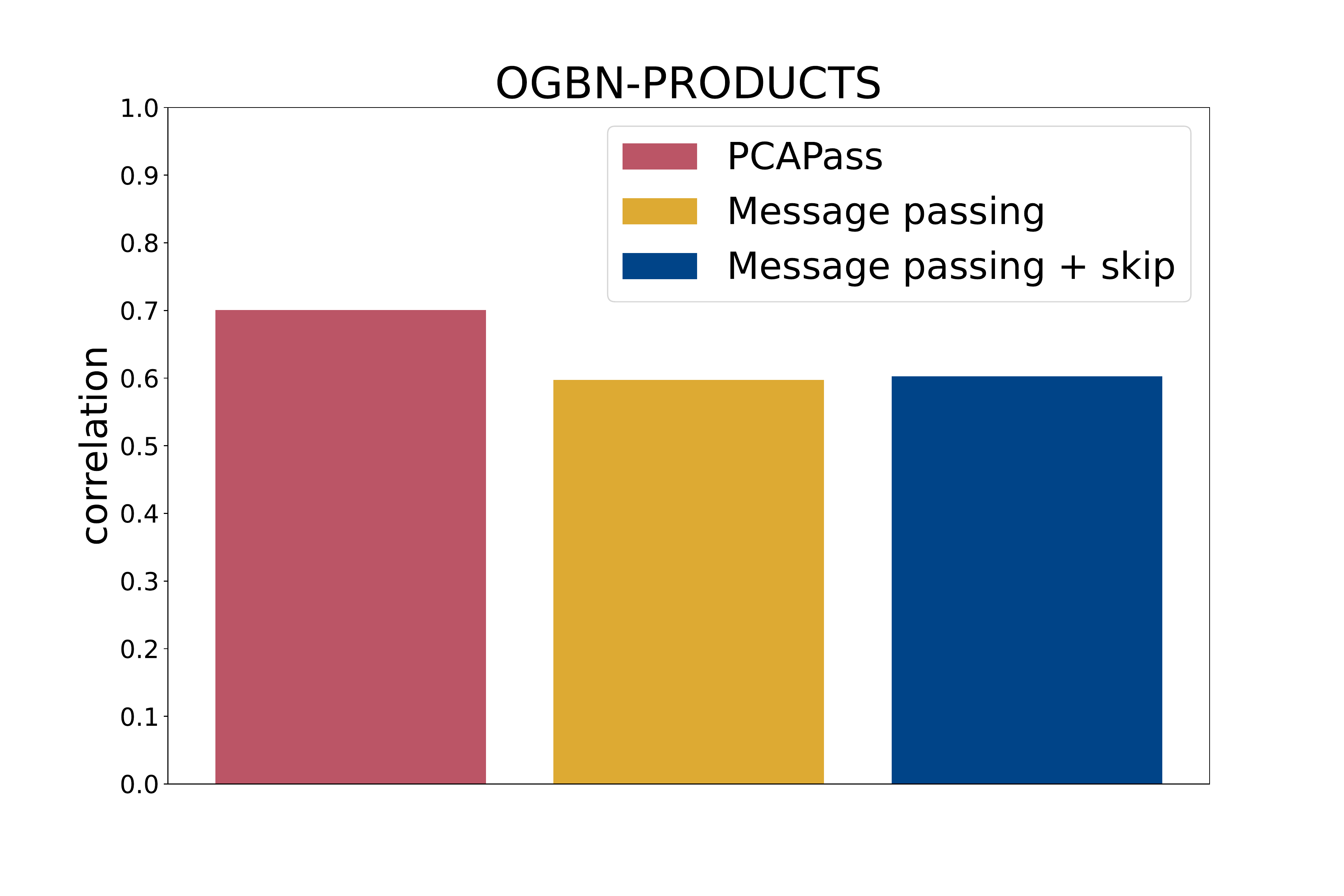}}
    \caption{Absolute value of the correlation between validation loss and test accuracy. We show that XGBoost, along with our node embedding method, best generalizes to unseen data.}
    \label{fig:pcapass_corellation}
    \end{center}
    \vskip -0.2in
\end{figure}

We perform additional analysis with 50 runs of hyperparameter optimization applied to both preprocessing and the XGBoost classifier (Figure \ref{fig:pcapass_validation}). 
To evaluate how solutions generalize we choose a validation cross-entropy loss to indicate model metric evaluated during optimization and test accuracy to show performance on unseen data. 
Our results show that classifier achieves the best results with PCAPass node embeddings and generalizes better than other methods due to the strongest Pearson correlation between evaluated metrics (Figure \ref{fig:pcapass_corellation}).
Interestingly, message passing and its variant with skip connections yield similar results, suggesting the latter may be suffering from over-squashing.

\subsection{Long-range Aggregation}
To test the resistance of our method to over-smoothing, we performed an analysis using the k-means clustering algorithm (Figure~\ref{fig:vms}). We chose this approach because it is a simple solution and has no hyperparameters that would introduce the bias in the results.
We explored PCAPass, message passing, and message passing with skip connections across the 50 $k$-hop neighborhood aggregation step. 
As a result of over-smoothing, the nodes become more similar to each other, which makes clustering more difficult.
We normalized the result of v-measure per method to show how resistant it is to this problem with respect to the size of the receptive field.
For each of the methods, we chose the mean as the aggregation function and we did not use the possibility of PCAPass to resize node embeddings to have a fair comparison.
Additionally, we standardize node embeddings for each of the methods before applying k-means. PCAPass and message passing with skip connections have similar trend lines on ogbn-products and ogbn-arxiv datasets which shows that PCA does not spoil the effect of skip connections.
Although our method performance drops faster on Reddit than message passing, it still has its best result with a larger receptive field.

\section{Conclusion and Future Work}
In this paper, we presented PCAPass, an approach for generating graph node embeddings. Our method uses skip connections and dimensionality reduction to retain information about the node and its neighborhood. The use of these techniques provides a mechanism for navigating over-smoothing and over-squashing, allowing information to be gathered from distant neighborhoods. 
PCAPass, which uses message passing during the preprocessing step, allows the GBDT classifier to use the graph topology efficiently.
Our method combined with XGBoost gives competitive results compared to popular GNNs, which we demonstrated on 3 benchmark datasets. 
We have empirically demonstrated that our solution creates embeddings that lead to better generalization than with message passing alone.
Additionally, PCAPass can be used by other GNNs that use message passing in preprocessing, such as SIGN, GAMLP and NARS, giving the possibility to aggregate information from distant neighborhoods.

We use mean aggregation in our experiments, but PCAPass can potentially benefit from using more specialized aggregation functions suited for graph structure and performed task. 
In addition, our architecture allows the use of other dimesionality reduction algorithms to create more informative embeddings.
Our results shows that dimensionality reduction could be a promising mechanism for alleviating over-squashing in the message passing process. 
It can be included in neighborhood aggregation mechanism to leave as much relevant information as possible.

\nocite{langley00}

\bibliography{pcapass}
\bibliographystyle{icml2022}

\end{document}